\documentclass{article} % For LaTeX2e
\usepackage{nips15submit_e,times}
\usepackage{hyperref}
\usepackage{url}

\usepackage{times}

\usepackage{amsmath}
\usepackage{amssymb}
\usepackage{epstopdf}
\usepackage{multirow}
\usepackage{colortbl}

\definecolor{Gray}{gray}{0.85}
\usepackage{algorithm}
\usepackage[noend]{algpseudocode}
\usepackage{graphicx}

\usepackage{booktabs}
\usepackage{bbold}

\usepackage{wrapfig}

\hypersetup{
    colorlinks=false,
    pdfborder={0.1 0.1 0.1},
}

\makeatletter
\newcommand*{\rom}[1]{\expandafter\@slowromancap\romannumeral #1@}
\makeatother

\title{Nonlinear Metric Learning for $k$NN and SVMs through Geometric Transformations}

\author{
Bibo Shi, \quad Jundong Liu  \\
School of Electrical Engineering and Computer Science\\
Ohio University\\
Athens, OH 45701 \\
\texttt{bs354409@ohio.edu, liu@cs.ohio.edu } \\
}

\nipsfinalcopy % Uncomment for camera-ready version

\begin{document}

\maketitle
\begin{abstract}

In recent years, research efforts to extend linear metric learning models to handle nonlinear structures have attracted great interests. In this paper, we propose a novel nonlinear solution through the utilization of deformable geometric models to learn spatially varying metrics, and apply the strategy to boost the performance of both $k$NN and SVM classifiers. 
Thin-plate splines (TPS) are chosen as the geometric model due to their remarkable versatility and representation power in accounting for high-order deformations. 
By transforming the input space through TPS, we can pull same-class neighbors closer while pushing different-class points farther away in $k$NN, as well as make the input data points more linearly separable in SVMs. 
%We have developed the formulations and solutions for both $k$NN and SVM classifiers in this paper. 
 Improvements in the performance of $k$NN classification are demonstrated through experiments on synthetic and real world datasets, with comparisons made  with several state-of-the-art metric learning solutions. Our SVM-based models also achieve significant improvements over traditional linear and kernel SVMs with the same datasets.

\end{abstract}

% In this paper, we present two nonlinear metric learning solutions, for $k$NN and SVMs respectively, based on geometric transformations. The novelty of our approaches lies in the fact that it generalizes the linear or piecewise linear transformations in traditional metric learning solutions to a globally smooth nonlinear deformation in the {\bf input} space. The geometric model used in this paper is thin-plate splines, and it can be extended to other radial distance functions. To explore other types of geometric models from the perspective of conditionally positive definite kernels is the direction of our future efforts. We are also interested in investigating a more efficient numerical optimization scheme (or the analytic form) for the proposed ML-TPS methods.

% Our solution is a direct generalization of linear metric learning through the application of deformable geometric models to transform the entire input space. 

%We have designed TPS-based ML solutions for both kNN and SVM classifiers, which will be presented in next two sections. To our best knowledge, this is the first work that utilizes nonlinear dense transformations, or spatially varying deformation models in metric learning. Our experimental results on synthetic and real data demonstrate the effectiveness of the proposed methods.

 %\vspace{-0.05in}
\section{Introduction}

%In mathematics, a metric is a function that defines the distance between a pair of data points. It plays a crucial role in many machine learning and data mining algorithms, e.g., $k$-means clustering and $k$NN classifier. Metric learning is to learn a distance function for the input space of data based on a given collection of similar/dissimilar constraints that preserve the distance relation among the training data pairs. General-purpose metrics (e.g., Euclidean, Manhattan, cosine, etc.) exist, but they often fail to capture the inherent characteristics of the data of interest. In contrast, a learned metric, tailored to the inputs, can significantly improve the algorithms' performance in many classification, clustering and retrieval tasks \cite{bellet2013survey,yang2006distance}.
Many machine learning and data mining algorithms rely on Euclidean metrics to compute pair-wise dissimilarities, which assign equal weight to each feature component. Replacing Euclidean metric with a learned one from the inputs can often significantly improve the performance of the algorithms \cite{bellet2013survey,yang2006distance}. Based on the form of the learned metric, metric learning (ML) algorithms can be categorized into linear and nonlinear groups \cite{yang2006distance}. Linear models \cite{xing2002,NCA,RCA,ITML,weinberger2009distance, MCML} commonly try to estimate a ``best'' affine transformation to deform the input space, such that the resulted Mahalanobis distance would very well agree with the supervisory information brought by training samples. Many early works have focused on linear methods as they are easy to use, convenient to optimize and less prone to overfitting \cite{bellet2013survey}. 
However, when handling data with nonlinear structures, linear models show inherently limited expressive power and separation capability --- highly nonlinear multi-class boundaries often can not be well modeled by a single Mahalanobis distance metric.

Generalizing linear models for nonlinear cases have gained steam in recent years, and such extensions have been pushed forward mainly along kernelization \cite{torresani2007kLMNN,kwok2003learning,chatpatanasiri2010new} and localization \cite{LMNN2009,msNCA,ramanan2011,noh2010generative,wang2012parametric} directions . % Nonlinear models are usually designed through kernelization or localization of certain linear models. 
The idea of kernelization \cite{torresani2007kLMNN,kwok2003learning} is to embed the input features into a higher dimensional space, with the hope that the transformed data would be more linearly separable under the new space. While kernelization may dramatically improve the performance of linear methods for many highly nonlinear problems, solutions in this group are prone to overfitting \cite{bellet2013survey}, and their utilization is inherently limited by the sizes of the kernel matrices \cite{he2013kernel}. Localization approaches focus on combining multiple local %piecewise 
metrics, which are learned based on either local neighborhoods or class memberships. The granularity levels of the neighborhoods vary from per-partition \cite{msNCA,ramanan2011}, per-class \cite{LMNN2009} to per-exemplar \cite{noh2010generative,wang2012parametric}. A different strategy is adopted in the GB-LMNN method \cite{kedem2012non}, which learns a global nonlinear mapping by iteratively adding nonlinear components onto a linear metric. At each iteration, a regression tree of depth $p$ splits the input space into $2^p$ axis-aligned regions, and points falling into the regions are shifted in different directions.
% utilizes an additive function $\phi = \phi_0 + \alpha \sum_{t=1}^T h_t$, where $h_t$ are gradient boosted regression trees of limited depth $p$. Although a global nonlinear transformation is learned, its underlying gradient update still benefits from the piecewise localization, where the input space is divided into $2^p$ regions by the selected tree at each gradient descent iteration, with samples in the same region translated in the same direction.
While the localization strategies 
%provides more power to deform the feature space, 
are usually more powerful in accommodating nonlinear structures, generalizing these methods to fit other classifiers than $k$NN is not trivial. 
To avoid non-symmetric metrics, extra cares are commonly needed to ensure the smoothness of the transformed input space. In addition, estimating geodesic distances and group statistics on such metric manifolds are often computationally expensive.

Most of the existing ML solutions are designed based on pairwise distances, and therefore best suited to improve nearest neighbor (NN) based algorithms, such as $k$-NN and $k$-means. Typically, a two-step procedure is involved: a best metric is first estimated  through training samples, followed by the application of the learned metric to the ensuing classification or clustering algorithms. Since learning a metric is equivalent to learn a feature transformation \cite{bellet2013survey}, metric learning can also be applied to SVM models, either as a preprocessing step \cite{xu2012distance}, or as an input space transformation \cite{xu2012distance,zhu2012learning,WuZZL15}. In \cite{xu2012distance}, Xu {\it et al.} studied both approaches and found applying linear transformations  to the input samples outperformed three state-of-the-art linear ML models utilized as preprocessing steps for SVMs. Several other transformation-based models \cite{zhu2012learning,WuZZL15} have also reported improved 
classification accuracies over the standard linear and kernel SVMs. However, all the models employ linear transformations, which limit their capabilities in dealing with complex data.
%Recently, some efforts have been done to combine metric learning with Support Vector Machines \cite{SVML,zhu2012learning}. 

In light of the aforementioned limitations and drawbacks of the existing models, we propose a new nonlinear remedy in this paper. Our solution is a direct generalization of linear metric learning through the application of deformable geometric models to transform the entire input space. 
% {\bf Supervisory constraints that preserve the distance relation among the training data pairs are optimized}, leading to improved classification performance. %The term deformable (as opposed to linear or global) is used to denote the fact that the observed signals are associated through a nonlinear dense transformation, or a spatially varying deformation model.
%\paragraph{Deformable models: from point registration perspective} 
In this study, we choose {\it thin-plate splines} (TPS) as the transformation model, and the choice is with the consideration of the compromise between computational efficiency and richness of description. TPS are well-known for their remarkable versatility and representation power in accounting for high-order deformations. We have designed TPS-based ML solutions for both $k$NN and SVM classifiers, which will be presented in next two sections. To our best knowledge, this is the first work that utilizes nonlinear dense transformations, or spatially varying deformation models in metric learning. Our experimental results on synthetic and real data demonstrate the effectiveness of the proposed methods.

\section{Nonlinear Metric Learning for Nearest Neighbor }

%The power of linear metric learning methods in handling data with nonlinear structures is inherently limited. 
 Many linear metric learning models are formulated under the nearest neighbor (NN) paradigm, with the same goal that the estimated transformation would pull similar data points closer while pushing dissimilar points apart. Our nonlinear ML model for NN is designed with the same idea. However, instead of using a single linear transformation, we choose to deform the input space nonlinearly through powerful radial basis functions -- thin-plate splines (TPS). With TPS, nonlinear metrics are computed globally, with smoothness ensured across the entire data space. Similarly as in linear models, the learned pairwise distance is simply the Euclidean distance after the nonlinear projection of the data through the estimated TPS transformation. 

 In this section, a pioneer Mahalanobis ML for clustering method (MMC) proposed by Xing {\it et al.}  \cite{xing2002} will be used as the platform to formulate our nonlinear ML solution for NN.
%under NN paradigm. 
Therefore, we will briefly review the concept of MMC first. Then we will describe the theoretical background of the TPS in the general context of transformations, 
% based on the mathematical work of \cite{duchon1977splines,wahba1990spline}, 
followed by the presentation of our proposed model.

\subsection{Linear Metric Learning and MMC} 

Given a set of training data instances $\mathcal{X} = \{\mathbf{x}_i| \ \mathbf{x}_i\in \mathbb{R}^{d}, i=1,\cdots, n\}$, where $n$ is the number of training samples, and $d$ is the number of features that a data instance has, the goal of ML is to learn a ``better'' metric function $D: \mathcal{X} \times \mathcal{X} \rightarrow \mathbb{R}$ to the problem of interest with the information carried by the training samples. % As mentioned before, 
Mahalanobis metric is one of the most popular metric functions used in existing ML algorithms \cite{NCA,RCA,MCML,weinberger2009distance,hoi2006DCA,msNCA}, which is defined by $D_M(\mathbf{x}_i,\mathbf{x}_j) = \sqrt{(\mathbf{x}_i - \mathbf{x}_j)^T M (\mathbf{x}_i-\mathbf{x}_j)}$. The control parameter $M  \in \mathbb{R}^{d \times d}$ is a square matrix. In order to qualify as a valid (pseudo-)metric, $M$ has to be positive semi-definite (PSD), denoted as $M\succeq 0$. As a PSD matrix, $M$ can be decomposed as $M = L^T L$, where $L \in \mathbb{R}^{k\times d}$ 
and $k$ is the rank of $M$. Then, $D_M(\mathbf{x}_i,\mathbf{x}_j)$ can be rewritten as follows:
%\vspace{-1mm}
\begin{equation} 
\label{eqn:linear}
\small
 D_M(\mathbf{x}_i,\mathbf{x}_j)  =  \sqrt{(\mathbf{x}_i - \mathbf{x}_j)^T L^T L(\mathbf{x}_i-\mathbf{x}_j)} =  \sqrt{(L \mathbf{x}_i - L \mathbf{x}_j)^T (L \mathbf{x}_i- L \mathbf{x}_j)}.
\end{equation}

Eqn.~\eqref{eqn:linear} explains why learning a Mahalanobis metric is equivalent to learning a linear transformation function and computing the Euclidean distance over the transformed data domain.

With the side information embedded in the class-equivalent constraints %$\mathcal P$ 
$\mathcal{P} = \{(\mathbf{x}_i,\mathbf{x}_j)| \ \mathbf{x}_i \textrm{ and }  \mathbf{x}_j \textrm{ belong to the same class} \}$ and class-nonequivalent constraints %$\mathcal N$
$\mathcal{N} = \{(\mathbf{x}_i,\mathbf{x}_j) \\ | \ \mathbf{x}_i  \textrm{ and }  \mathbf{x}_j \textrm{ belong to different classes} \}$, MMC model formulates the problem of ML into the following convex programming problem: 

%\vspace{-1mm}
\begin{equation}
\small
 %\begin{aligned}
\underset{M}{\text{min}} \quad  J(M) = \sum \limits_{\mathbf{x}_i, \mathbf{x}_j \in \mathcal P} D_{M}^{2}(\mathbf{x}_i, \mathbf{x}_j)  \quad  \text{  s.t. } \quad  M \succeq 0,  \quad \sum \limits_{\mathbf{x}_i, \mathbf{x}_j \in \mathcal N} D_{M}^{2}(\mathbf{x}_i, \mathbf{x}_j) \geq 1 .\\
 %\end{aligned}
\end{equation}

The objective function aims at improving the subsequent NN based algorithms via minimizing the sum of distances between similar training data, while keeping the sum of distances between dissimilar ones large. Note that, besides the PSD constraint on $M$, an additional constraint on the training samples in $\mathcal N$ is needed to avoid trivial solutions for the optimization. To solve this optimization problem, the projected gradient descent method is used, which projects the estimated matrix back to the PSD group whenever it is necessary. %This numerical procedure involves an eigenvalue decomposition at each iteration, and as a result, the computation cost is a nonnegligible issue when dealing with high dimensional data. 
%Later in 2003, this algorithm is extended to be nonlinear by the introduction  of kernelization \cite{kwok2003_kernelxing}. \\

%\paragraph{other linear solutions: MMC, RCA, LMNN}

%\paragraph{Nonlinear extension: msNCA, mm-LMAA, kernel LMNN and more}

%\begin{document}
% {\it Objective: This short summary describes how to extend the current metric learning methods to be nonlinear through TPS. It contains the development of our models: from simple ML-TPS, to constrained ML-TPS, and to Diffeo-ML-TPS.} 

%\subsection{Nonlinear Metric Learning through TPS (NN-ML-TPS)}

\subsection{TPS}

Thin-plate splines (TPS) are the high-dimensional analogs of the cubic splines in one dimension, and have been widely used as an interpolation tool in the research of data approximation, surface reconstruction, shape alignments, etc. When it is utilized to align a set of $n$ corresponding point-pairs $\mathbf{u}_i$ and $\mathbf{v}_i,$ ($i=1,\dots,n$), a TPS transformation is a mapping function $f(\mathbf{x}):\mathbb{R}^{d} \rightarrow \mathbb{R}^{d}$ within a suitable Hilbert space $\mathcal{H}$, that matches $\mathbf{u}_i$ and $\mathbf{v}_i$, as well as minimizes a smoothness TPS penalty functional $J^{d}_{m}(f):\mathcal{H} \rightarrow \mathbb{R}$ (will be given in Eqn.~\ref{TPS_I}).

Typically, the problem of finding $f$ can be decomposed into $\mathnormal{d}$ interpolation problems, finding component thin plate splines $f_k, k=1, \dots, \mathnormal{d}$, separately. Suppose the unknown interpolation function $f_k:\mathbb{R}^{d} \rightarrow \mathbb{R}$ belongs to the Sobolev space $\mathcal{H}^{m}(\mathnormal{\Omega)}$, where $\mathnormal{m}$ is an unknown positive integer and $\mathnormal{\Omega}$ is an open subset of $\mathbb{R}^{d}$, TPS transformations minimize the smoothness penalty functional of the following general form:
%\vspace{-0.05in}
\begin{equation}
\small
\begin{aligned}
J^{d}_{m}(f) =  \int ||\mathcal{D}^{m} f||^{2} \mathrm{d}X 
  = \sum_{a_1+\dots+a_d=m} \frac{m!}{a_1!\dots a_d!} \int \dots \int (\frac{\partial^{m}f}{\partial{x}^{a_1}_1\dots\partial{x}^{a_d}_d})^2 \mathrm{d}x_1\dots \mathrm{d}x_d  \\
% &J(f)& = & \quad J^{d}_{m}(f) \ \ \ \ = \ \ \ \ \ \int ||\mathcal{D}^{m} f(X))||^{2} \mathrm{d}X \\
%  &              & = & \sum_{a_1+\dots+a_d=m} \frac{m!}{a_1!\dots a_d!} \int \dots \int (\frac{\partial^{m}f}{\partial{x}^{a_1}_1\dots\partial{x}^{a_d}_d})^2 \mathrm{d}x_1\dots \mathrm{d}x_d  \\
\end{aligned}
\label{TPS_I}
\end{equation}
where $\mathcal{D}^{m}f$ is the matrix of $m$-th order partial derivatives of $f$, with $\mathnormal{a}_k$ being positive, and $\mathrm{d}X = \mathrm{d}x_1 ... \mathrm{d}x_d$, where $x_j$ are the components of $\mathbf{x}$.  The penalty functional is the generalized form for the space integral of the squared second order derivatives of the mapping function.  We will suppose the mapping function $f \in \mathcal{C}$, a space of functions whose partial derivatives of total order $m$ are in $\mathcal{L}_2(\mathbb{R}^d)$. To have the evaluation functionals bounded in $\mathcal{C}$, we need $\mathcal{C}$ to be a reproducing kernel Hilbert space (r.k.h.s.), endowed with the seminorm $J^{d}_{m}(f)$. For this, it is necessary and sufficient that $2\mathnormal{m-d}>0$. 
The null space of $J^{d}_{m}(f)$ consists of a set of polynomial functions $\phi_m$ with maximum degree of $(\mathnormal{m}-1)$, so the dimension of this null space is $\mathnormal{N}_0 = (d+m-1)!/(d!(m-1)!)$.

The main problem of TPS is that $\mathnormal{N}_0$, the dimension of the null space, increases exponentially with $d$ due to the requirement of $2\mathnormal{m-d}>0$. To solve this problem, Duchon \cite{duchon1977splines} proposed to replace $J^{d}_{m}(f)$ by its weighted squared norm in Fourier space. Since the Fourier transform, denoted as $\mathcal{F}(.)$ is isometric, the penalty functional $J^{d}_{m}(f)$ can be replaced by its squared norm in Fourier space: 
%\vspace{-0.08in}
\begin{equation}
\small
\int ||\mathcal{D}^{m} f(t))||^{2} \mathrm{d}X  \iff \int  \mathcal{F}(\mathcal{D}^{m} f(\tau))||^{2} \mathrm{d}\tau
\label{TPS_duchon}
\end{equation}

By adding a weighting function, Duchon introduced a new penalty functional to solve the exponential growth problem of the dimension for TPS' null space, which is defined as 
%\vspace{-0.04in}
\begin{equation}
\small
J^{d}_{m,s}(f) = \int |\tau|^{2s}|| \mathcal{F}(\mathcal{D}^{m} f(\tau))||^{2} \mathrm{d}\tau ,
\label{TPS_duchon}
\end{equation}
 provided that $2(m+s)-d>0$ and $2s<d$. As suggested by \cite{duchon1977splines}, one can select an appropriate $s$ to have a lower dimension for the null space of $J^{d}_{m,s}(f)$, with the maximum degree of the polynomial functions $\phi_{m,s}$ spanned in this null space  being decreased to $1$.

 The classic solution of  Eqn. (\ref{TPS_duchon}) has a representation in terms of a radial basis function (TPS interpolation function), 
%$f(x) = \sum_{i=1}^n a_i G(||{\bf x}-{\bf x}_i||)+b{\bf x}+c$, 
\begin{equation}
f_k(\mathbf{x}) = \sum_{i=1}^n \psi_i G(||\mathbf{x}-\mathbf{x}_i||)+ \boldsymbol{\ell}^T\mathbf{x} + c ,
\label{tps_interpolation}
\end{equation}

where $||.||$ denotes the Euclidean norm and $\{\psi_i\}$ are a set of weights for the nonlinear part; $\boldsymbol{\ell}$ and $c$ are the weights for the linear part. The corresponding radial distance kernel of TPS, which is the Green's function to solve Eqn. (\ref{TPS_duchon}), is as follows:    
\begin{eqnarray}
\small 
G({\bf x},{\bf x}_i) =  \ G(||{\bf x}-{\bf x}_i||)
\propto  \begin{cases}  ||{\bf x}-{\bf x}_i||^{2(m+s)-d} \mathrm{ln}||{\bf x}-{\bf x}_i||,  \quad \text{\small{if  $2(m+s)-d$ is even; }} \\ 
                                                     %& \text{\small{integer; }}\\                                          
                     ||{\bf x}-{\bf x}_i||^{2(m+s)-d}, \qquad \qquad \quad \mbox{otherwise.}    
            \end{cases}  %\nonumber 
\end{eqnarray}

% 
% The classic solution for minimizing Eqn. (\ref{TPS_duchon}) has a representation in terms of radial basis function, 
% \begin{equation}
% f_k(\mathbf{x}) = \sum_{j=1}^{N_0} b_j \phi_j(x) + \sum_{i=1}^n w_i G(||x-\mathbf{p}_i||) 
% \label{TPS_fk}
%  \end{equation}
% 
% $||.||$ denotes Euclidean norm; and $\{b_j\}$ are a set of weights for linear part; $\{w_i\}$ are the weights for nonlinear part. The corresponding radial distance kernel of TPS, $G(x,x_k)$, is as follows:    
% \begin{equation}
% \small
% \begin{aligned}
% &G(x,x_k)& = & G(||x-x_k||)\\
% &  & \propto & \begin{cases} |x-x_k|^{2} ln|x-x_k|, & \text{\small{if  $2v+2s-d$ is even; }} \\ 
%                                                      %& \text{\small{integer; }}\\                                          
%                      |x-x_k|, & \mbox{otherwise.}
%             \end{cases}
% \end{aligned}
% \end{equation}
% 
% 

For more details about TPS, we refer readers to \cite{duchon1977splines, wahba1990spline}. 

% The Thin Plate Spline has a natural representation in terms of radial basis functions. Given a set of control points \{w_{i}, i = 1,2, \ldots,K\}, a radial basis function basically defines a spatial mapping which maps any location x in space to a new location f(x), represented by,
%Just as for the cubic spline, $f$ need only have fourth-order partial derivatives that are contineous almost everywhere except at the data points. 
\subsection{TPS Metric Learning for Nearest Neighbor (TML-NN)}
The TPS transformation for point interpolation, as specified in Eqn.~(\ref{tps_interpolation}), can be employed as the geometric model to deform the input space for nonlinear metric learning. Such a transformation would ensure certain desired smoothness as it minimizes the bending energy $J^{d}_{m}(f)$ in Eqn.~(\ref{TPS_I}). Within the metric learning setting, let $\mathbf{x}$ be one of the training samples in the original feature space $\mathcal{X}$ of $d$ dimensions, and $f(\mathbf{x})$ be the transformed destination of $\mathbf{x}$, also of $d$ dimensions. %{\bf $x_l$ is the $l$th component of variable $x$.}
 Through a straightforward mathematical manipulations \cite{ChuiCVIU03}, we can get $f(\mathbf{x})$ in matrix format:  
% The previous $f(x)$ is for interpolation, which maps a vector $x$ to a scalor. Instead, we should use the tps $f(x)$ describes in Haili Chui's paper:
\begin{equation}
\small
f(\mathbf{x}) = L  \mathbf{x}   + \Psi  \begin{pmatrix} G(\mathbf{x}, \mathbf{x}_1) \\ \cdots \\ G(\mathbf{x}, \mathbf{x}_p) \\  \end{pmatrix} = L  \mathbf{x}  + \Psi \vec{G}(\mathbf{x}) ,
%f(\vec{X}) = \vec{X} \cdot B  + \begin{pmatrix} G(x_1, x_1) \ G(x_2, x_1)... G(x_n, x_1) \\ G(x_1, x_2) \ G(x_2, x_2)... G(x_n, x_2) \\ \cdots \\ G(x_1, x_n) \ G(x_2, x_n)... G(x_n, x_n) \\  \end{pmatrix} \cdot W = \vec{X} \cdot B + \vec{G} \cdot W
\label{f_in_TPS}
\end{equation}
where $L$ (size $d\times d$) is a linear transformation matrix, corresponding to $L$ in  Mahalanobis  metric, %and 
$\Psi$ (size $d\times p$) is the weight matrix for the nonlinear parts, and $p$ is the number of anchor points ($\mathbf{x}_1, \dots, \mathbf{x}_p$) to compute the TPS kernel.   
%$\mathbf{x}_p$ are the anchor points used to compute the TPS kernel. 
Usually, we can use all the training data points as the anchor points. However, in practice, $p$ anchor points are extracted via different methods to describe the whole input space under the consideration of computational cost, such as k-medoids method used in \cite{wang2012parametric}.   
%{\bf Overall, we have only $n+m$ (or $n+m^2$) variables to optimize: $w_k$ and $b_l$.} Note: $b_0$ is for translation, which is useless in metric learning ({\bf but you didn't even mention what $b_0$ is}). 

%The first section describes our original effort to combine Metric Learning (ML) and TPS.

%\subsection{The simplest ML objective}

%To simplify the demenstration of the nonlinear extension, we directly adopt the Eric Xing's objective with minor changes. Given a set of training data instances $X= \{x_i|x_i\in \mathbf{R}^{m}, i=1,\cdots, n\}$, where $n$ is the number of training samples, and $m$ is the number of features that a data instance has. The following are the constraint forms used here:

The goal of our ML solution is also pulling the samples of the same class closer to each other while pushing different classes further away, directly through a TPS nonlinear transformation as described in Eqn.~(\ref{f_in_TPS}). This can be achieved through the following constrained optimization: 
\vspace{-0.02in}
\begin{equation}
\small
 \begin{aligned}
 &\underset{L, \Psi}{\text{min}}
 & & J =  \sum \limits_{\mathbf{x}_i, \mathbf{x}_j \in \mathcal P}  \lVert f(\mathbf{x}_i)-f(\mathbf{x}_j)\rVert^2 + \lambda \lVert \Psi \rVert^2_{F}   \\
 &\text{s.t.} 
 && \sum \limits_{\mathbf{x}_i, \mathbf{x}_j \in \mathcal N}   \lVert f(\mathbf{x}_i)-f(\mathbf{x}_j)\rVert^2 \geq 1; \quad  \sum \limits_{i=1}^p  \Psi_i^k =0, \quad \sum \limits_{i=1}^p \Psi_i^k \mathbf{x}_i^k =0, \text{ }\forall k = 1 \dots d.
 \end{aligned}
 \label{ml_tps_con}
\end{equation}

$f$ is in the form of Eqn.~(\ref{f_in_TPS}); $\Psi^k$ is the $k$th column of $\Psi$; $\mathbf{x}^k$ is the $k$th component of $\mathbf{x}$. 
Compared with MMC, another component $\tiny \lVert \Psi \rVert^2_{F}$, the squared Frobenius norm of $\Psi$, is added to the objective function as a regularizer to prevent  overfitting. $\lambda$ is the weighting factor to control the importance of two components. 
Similarly as in MMC, the nonequivalent constraint {$\tiny{\sum_{\mathbf{x}_i, \mathbf{x}_j \in \mathcal N}   \lVert f(\mathbf{x}_i)-f(\mathbf{x}_j)\rVert^2 \geq 1}$} is to impose a scaling control to avoid trivial solutions.
The other two equivalent constraints with respect to $\Psi$ %(the last line of Eqn.~(\ref{ml_tps_con}))  
% :
% \vspace{-2mm}
%  \begin{equation} 
%   %\small
%    \forall k = 1 \dots m: \quad \sum \limits_{i=1}^n W_i^k =0, \quad \sum \limits_{i=1}^n W_i^k x_i^k =0.
% \end{equation}
is to ensure that the elastic part of the transformation is zero at infinity  \cite{rohr2001landmark}.  

Due to the nonlinearity of TPS, it is difficult to analytically solve this nonlinear constrained problem. Alternatively, we can %convert the constrained problem into an unconstrained optimization, and 
use a gradient based constrained optimization solver \footnote{We use a SQP based constrained optimizer ``fmincon'' in Matlab Optimization Toolbox.} %to find the local minimum.}
to get a local minimum for Eqn.~(\ref{ml_tps_con}) . 
The complexity of our TML-NN model is dominated by the computation of the TPS kernel, which is $O(p*n^2)$, as well as the rate of convergence of the chosen gradient based optimizer. $n$ is the number of training samples, and $p$ is the number of anchor points. %extracted by  k-medoids, which should be at least larger than $d$, the dimension of the data points. %If the number of the anchor points is set to $n$, the total number of the training instances, then the time complexity of computing the TPS kernels will be $O(n^2)$. The other steps for training only operate on constant numbers, so their corresponding time can be ignored. 
%\vspace{-0.08in}

\begin{wrapfigure}{r}{0.66\textwidth}
 %\vspace{-0.15in}
  \begin{tabular}{cc}
 \includegraphics[width=0.32\columnwidth]{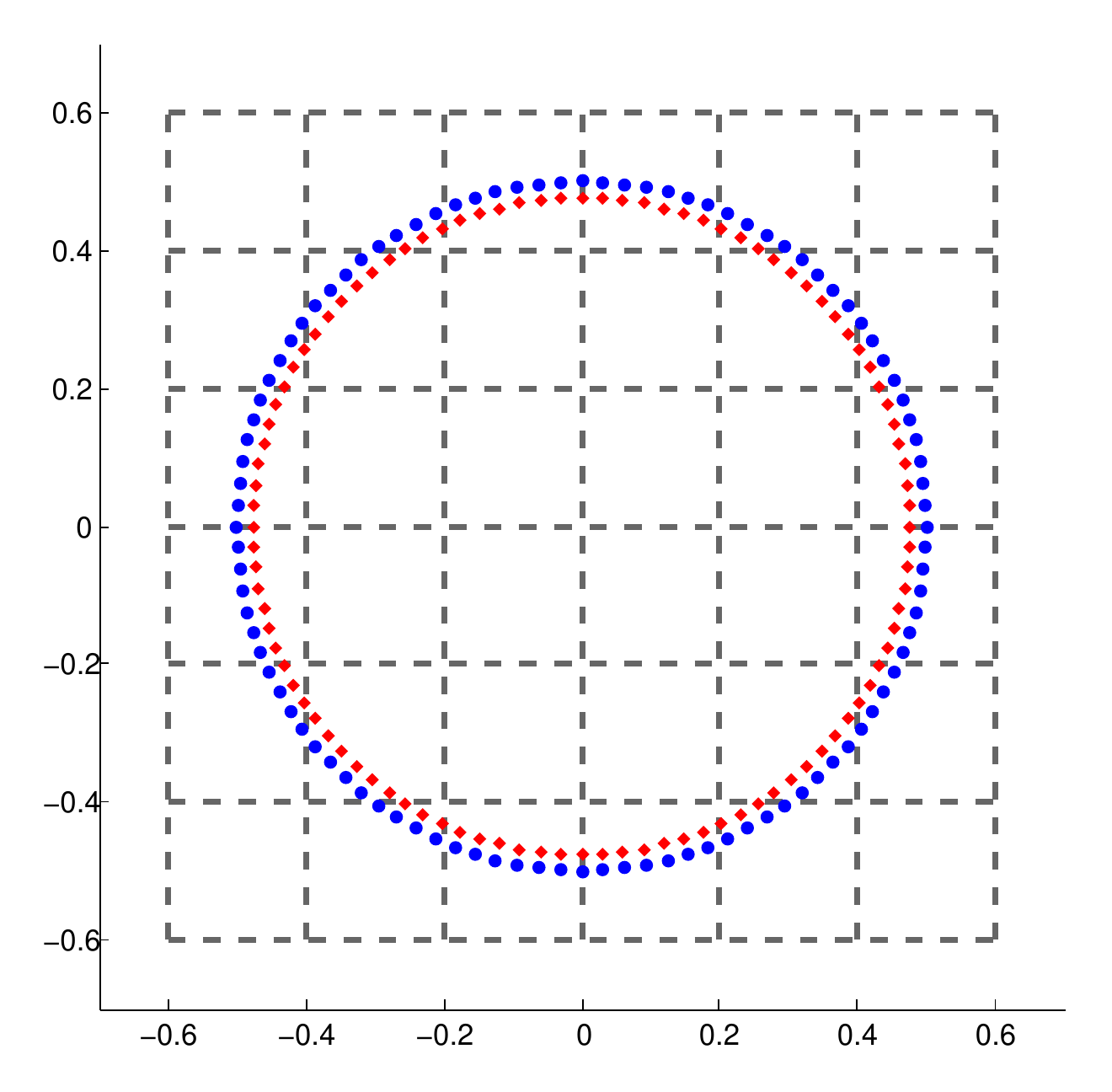} &
 \includegraphics[width=0.32\columnwidth]{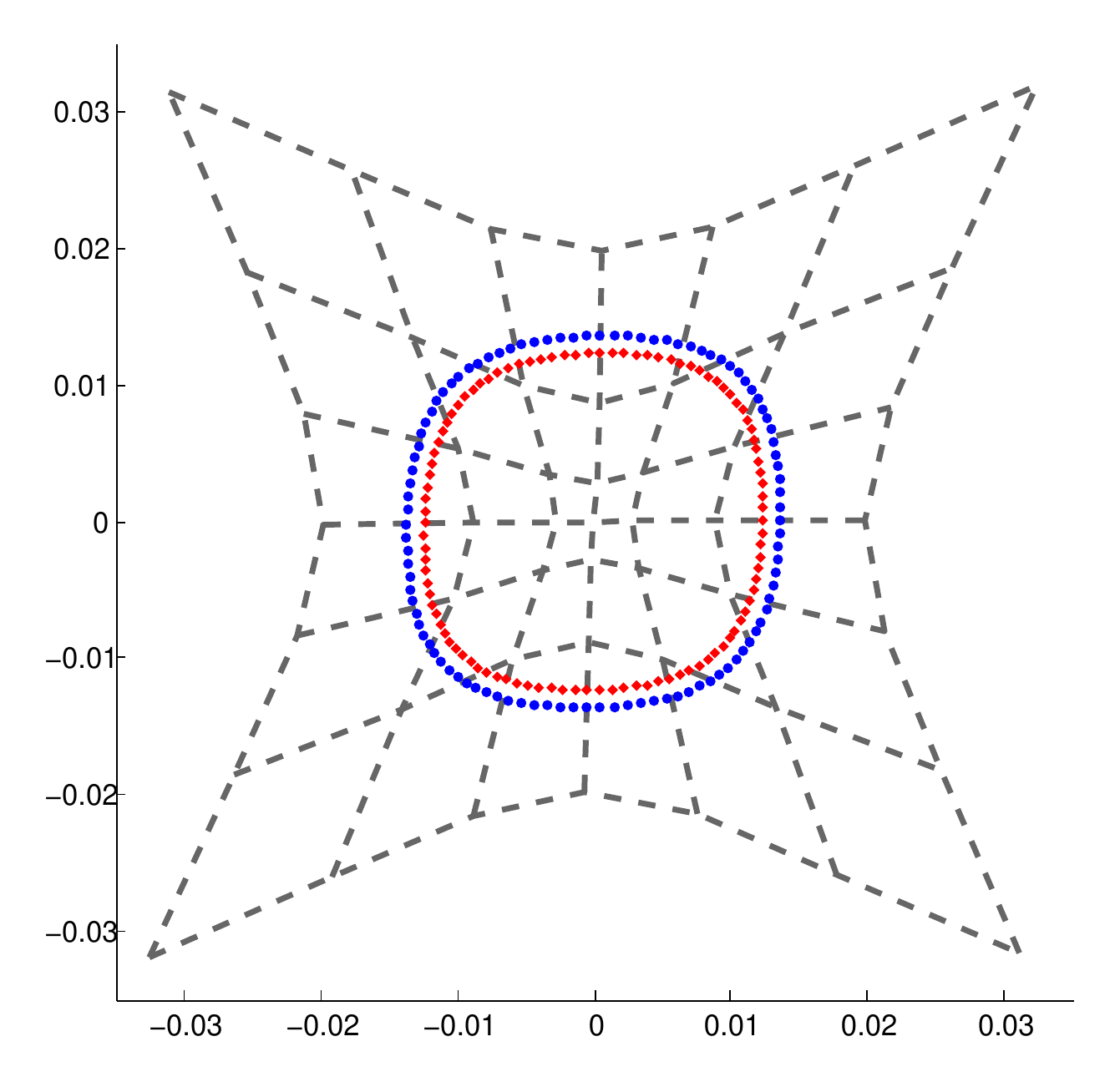} \\
 (a) & (b) \\
\end{tabular}
 %\vspace{-0.15in}
 \caption{(a) original inputs with coordinate grids; (b) transformed data and the deformation field generated by TML-NN.}
\label{fig:bibo_syn_exp}
\end{wrapfigure}
  To demonstrate the ability of TML-NN in handling nonlinear cases, we conducted a similar experiment used in the GB-LMNN method \cite{kedem2012non}. Fig.~\ref{fig:bibo_syn_exp}.(a) shows a synthetic dataset consisting of inputs sampled from two concentric circles (in blue dots and red diamonds), each of which defines a different class membership. Global linear transformations in linear metric learning are not sufficient to improve the accuracy of $k$NN ($k=1$) classification on this data set. As contrast, by utilizing TPS to model the underlying nonlinear transformation, as shown in Fig.~\ref{fig:bibo_syn_exp}.(b), we can easily enlarge the separation between outer and inner circles,  leading to improved classification rate (would be $100\%$ for $1$NN).

\section{TPS Metric Learning for Support Vector Machines (TML-SVM)}
 In this section, we present how to generalize our TPS metric learning model for SVMs. Similar as in \cite{zhu2012learning}, we formulate our model under the {\it Margin-Radius-Ratio} bounded SVM paradigm, which generalizes the traditional SVMs by bounding the estimation error \cite{weston2000feature}. Given training dataset $\mathcal{X} = \{\mathbf{x}_i| \ \mathbf{x}_i\in \mathbb{R}^{d}, i=1,\cdots, n\}$ together with the class label information $y_i \in \{-1,+1\}$, our proposed TML-SVM aims to simultaneously learn the nonlinear transformation as described in Eqn.~(\ref{f_in_TPS}) and a SVM classifier, which can be formulated as follows:
\vspace{-0.03in}
\begin{equation}
\small
 \begin{aligned}
 &\underset{L, \Psi, \mathbf{w}, b}{\text{min}}
 & & J =  \frac{1}{2} \lVert \mathbf{w} \rVert^2 + C_1 \sum_{i=1}^{n} \xi_i + C_2  \lVert \Psi \rVert^2_{F}   \\
 &\text{s.t.} 
 && y_i (\mathbf{w}^T f(\mathbf{x}_i) + b ) = y_i (\mathbf{w}^T (L\mathbf{x}_i+ \Psi \vec{G}(\mathbf{x}_i)) + b ) \geq 1- \xi_i, \quad \forall i = 1 \dots n  ; \text{ (\rom{1})}  \\
 &&& \xi_i \geq 0, \quad \forall i = 1 \dots n  ; \text{ (\rom{2})} \\
 &&& \lVert f( \mathbf{x}_i) - \mathbf{x}_c \rVert^2 = \lVert L\mathbf{x}_i+ \Psi \vec{G}(\mathbf{x}_i) - \mathbf{x}_c \rVert^2 \leq 1 , \quad \forall i = 1 \dots n  ; \text{ (\rom{3})} \\
 &&& \sum \limits_{i=1}^p  \Psi_i^k =0, \quad \sum \limits_{i=1}^p \Psi_i^k \mathbf{x}_i^k =0, \text{ }\forall k = 1 \dots d.   \text{ (\rom{4})}
 \end{aligned}
 \vspace{-0.03in}
 \label{ml_tps_SVM}
\end{equation}

The objective function combines the regularizer w.r.t. $\Psi$ for TPS transformation with the traditional soft margin SVMs. $C_1$ and $C_2$ are two trade-off hyper-parameters. The first two nonequivalent constraints (\rom{1} and \rom{2}) are the same as used in traditional SVMs. The third nonequivalent constraint (\rom{3}) is a unit-enclosing-ball constraint, which forces the radius of minimum-enclosing-ball to be unit in the transformed space and avoids trivial solutions. $\mathbf{x}_c$ is the center of all samples. In practice, we can simplify the unit-enclosing-ball constraint to $\lVert f( \mathbf{x}_i) \rVert^2 \leq 1$ through a preprocessing step to centralize the input data: $ \mathbf{x}_i \gets \mathbf{x}_i - \frac{1}{n}\sum_{i=1}^{n} \mathbf{x}_i$. The last two equivalent constraints are used to maintain the properties for TPS transformation at infinity, similar as in Eqn.~(\ref{ml_tps_con}).

To solve this optimization problem, we propose an efficient EM-like iterative minimization algorithm by updating $\{\mathbf{w},b\}$ and $\{L,\Psi\}$ alternatively. With $\{L,\Psi\}$ fixed, $f(\mathbf{x}_i)$ is explicit, and Eqn.~(\ref{ml_tps_SVM}) can be reformulated as:   
\vspace{-0.04in}
\begin{equation}
\vspace{-0.1in}
\small
 \begin{aligned}
  \underset{\mathbf{w}, b}{\text{min}} \quad J =  \frac{1}{2} \lVert \mathbf{w} \rVert^2 + C_1 \sum_{i=1}^{n} \xi_i \quad \text{s.t.} \quad 
  y_i (\mathbf{w}^T f(\mathbf{x}_i) + b ) \geq 1- \xi_i, \quad   \xi_i \geq 0, \quad \forall i = 1 \dots n  . \\
%  &\underset{w, b}{\text{min}}
%  & & J =  \frac{1}{2} \lVert \mathbf{w} \rVert^2_{2} + C_1 \sum_{i=1}^{n} \xi_i\\  % + C_2  \lVert T \rVert^2_{F}   \\
%  &\text{s.t.} 
%  && y_i (\mathbf{w}^T f(\mathbf{x}_i) + b ) \geq 1- \xi_i, \quad \forall i = 1 \dots n  ;   \\
%  &&& \xi_i \geq 0, \quad \forall i = 1 \dots n  . \\
 \end{aligned}
 \label{ml_tps_SVM_1}
\end{equation}
This becomes exactly the primal form of soft margin SVMs, which can be solved by off-the-shelf SVM solvers. With $\{\mathbf{w},b\}$ fixed, Eqn.~(\ref{ml_tps_SVM}) can be reformulated as: 
\vspace{-0.02in}
\begin{equation}
\small
 \begin{aligned}
 &\underset{L, \Psi}{\text{min}}
 & & J =   C_1 \sum_{i=1}^{n} \xi_i + C_2  \lVert \Psi \rVert^2_{F}   \\
 &\text{s.t.} 
 && y_i (\mathbf{w}^T f(\mathbf{x}_i) + b ) \geq 1- \xi_i, \quad \xi_i \geq 0, \quad \forall i = 1 \dots n  ;   \\
 &&& \lVert f( \mathbf{x}_i) \rVert^2  \leq 1 , \quad \forall i = 1 \dots n  ; \qquad \sum \limits_{i=1}^p  \Psi_i^k =0, \quad \sum \limits_{i=1}^p \Psi_i^k \mathbf{x}_i^k =0, \text{ }\forall k = 1 \dots d. \\
   \end{aligned}
 \label{ml_tps_SVM_2}
\end{equation}
By using hinge loss function, we can eliminate variables $\xi_i$, and reformulate Eqn.~(\ref{ml_tps_SVM_2}) as: 
\vspace{-0.02in}
\begin{equation}
\small
 \begin{aligned}
 &\underset{L, \Psi}{\text{min}}
 & & J =   C_1 \sum_{i=1}^{n} \text{max} [0, 1- y_i (\mathbf{w}^T f(\mathbf{x}_i) + b )]^2 + C_2  \lVert \Psi \rVert^2_{F}   \\
 &\text{s.t.} 
 && \lVert f( \mathbf{x}_i) \rVert^2  \leq 1 , \quad \forall i = 1 \dots n  ; \qquad \sum \limits_{i=1}^p  \Psi_i^k =0, \quad \sum \limits_{i=1}^p \Psi_i^k \mathbf{x}_i^k =0, \text{ }\forall k = 1 \dots d. \\
 \end{aligned}
 \label{ml_tps_SVM_3}
\end{equation}

As the squared hinge loss function is differentiable, it is not difficult to differentiate the objective function w.r.t $L$ and $\Psi$. Similarly as in solving Eqn.~(\ref{ml_tps_con}), we can also use a gradient based optimizer to get a local minimum for Eqn.~(\ref{ml_tps_SVM_3}), with the gradient computed as:
\vspace{-0.02in}
\begin{equation}
\small
 \begin{aligned}
 &\frac{\partial J}{\partial \Psi} & = &  - 2 C_1 \Psi \sum_{i=1}^{n} \text{max} [0, 1- y_i (\mathbf{w}^T f(\mathbf{x}_i) + b )](y_i \mathbf{w} \vec{G}^T(\mathbf{x}_i)) +  2C_2 \Psi \\
 & \frac{\partial J}{\partial L} & = &  - 2 C_1 \Psi \sum_{i=1}^{n} \text{max} [0, 1- y_i (\mathbf{w}^T f(\mathbf{x}_i) + b )](y_i \mathbf{w} \mathbf{x}_i^T) \\
 \end{aligned}
 \label{ml_tps_SVM_gradient}
\end{equation}
 To sum it up, the optimal nonlinear transformation defined by $\{L,\Psi\}$ along with the optimal SVM classifier coefficients $\{\mathbf{w},b\}$ can be obtained by an EM-like iterative procedure, as described in Algorithm~\ref{alg:diffeo}.

\begin{algorithm}
\caption{TPS Metric Learning for SVM (TML-SVM)}
\label{alg:diffeo}
\begin{algorithmic}[0]
\State \textbf{Input:} training dataset $\mathcal{X} = \{\mathbf{x}_i| \ \mathbf{x}_i\in \mathbb{R}^{d}, i=1,\cdots, n\}$, \\ 
\qquad \quad  class label information $y_i \in \{-1,+1\}$
\State \textbf{Initialize:} $\Psi = \mathbf{0}$, $L = \mathbf{I}$ 
\vspace{-0.04in}
\\\dotfill
\State \textbf{Centralize the input data:} $\mathbf{x}_i \gets \mathbf{x}_i - \frac{1}{n}\sum_{i=1}^{n} \mathbf{x}_i$
\State \textbf{Iterate the following two steps: } 
\State \quad  \textbf{(1) Update } $\{\mathbf{w},b\}$ \textbf{ with fixed } $\{L,\Psi\}$ \textbf{:}
\State \quad \quad \quad Compute the transformed data $f(\mathbf{x}_i)$ by following Eqn.~(\ref{f_in_TPS}) 
\State \quad \quad \quad Update $\{\mathbf{w},b\}$ by using off-the-shelf SVM solver with input of $f(\mathbf{x}_i)$
\State \quad  \textbf{(2) Update } $\{L,\Psi\}$ \textbf{ with fixed } $\{\mathbf{w},b\}$ \textbf{:}
\State \quad \quad \quad Update $\{L,\Psi\}$ by solving Eqn.~(\ref{ml_tps_SVM_3}) through gradient based optimizers \footnotemark 
%\State \quad \quad \quad Update $\{\mathbf{w},b\}$ by using off-the-shelf SVM solver with input of $f(\mathbf{x}_i)$
\State \textbf{ until convergence}
\vspace{-0.04in}
\\\dotfill
\State \textbf{Output: } the optimal SVM classifier defined by $\{\mathbf{w},b\}$,\\ \qquad \quad \quad the nonlinear TPS transformation defined by $\{L,\Psi\}$

\end{algorithmic}
\end{algorithm}
 \footnotetext{We still use ``fmincon'' in Matlab to solve Eqn.~(\ref{ml_tps_SVM_3}). In practice, the convergence for the second inner step is not necessary, so we use an early stop strategy to speed up the whole algorithm.} 

 \subsection{Kernelization of TML-SVM}
 
TML-SVM can be kernelized through a kernel principal component analysis (KPCA) based framework, as introduced in \cite{zhang2010general,chatpatanasiri2010new}. Unlike the traditional kernel trick \cite{scholkopf2001learning}, which often involves the derivation of new mathematical formulas, KPCA based framework provides an alternative choice that can directly utilize the original linear models. Typically, it consists of two simple stages: first, map the input data into a kernel feature space introduced by KPCA; then, train the linear model in this kernel space. Proved to be equivalent to the traditional kernel trick, this KPCA based framework also provides a convenient way to speed up a learner, if a low-rank KPCA is used. Through this procedure, kernelized TML-SVM can be easily realized by directly utilizing Algorithm~\ref{alg:diffeo} in the mapped KPCA space. For more details about this KPCA-based approach, we refer readers to  \cite{zhang2010general,chatpatanasiri2010new}.

%Let the selected kernel function be $K(\mathbf{x}_i,\mathbf{x}_j) = \phi(\mathbf{x}_i^T) \phi(\mathbf{x}_j^)$, where $\phi(\mathbf{x})$ defines an implicit mapping from the input data space to higher or infinite dimensional feature space. We can use KPCA to explictly express this mapping by using

\section{Experimental Results}

In this section, we present evaluation and comparison results of applying our proposed TPS-based nonlinear ML methods on seven widely used datasets from UCI machine learning repository. The details of these datasets are summarized in the leftmost column of Table~\ref{T:tatble}. The three numbers inside the bracket indicate data size, feature dimension, and number of classes for the corresponding dataset.  All datasets have been preprocessed through normalization.
%we present experimental results of evaluating our prposed TPS-based nonlinear metric learning methods on seven widely used datasets from UCI machine learning repository. 
To demonstrate the effectiveness of our proposed nonlinear metric learning method, we firstly choose $k$NN method as the baseline classifier, and compare the improvements made by TML-NN against five state-of-the-art NN based metric learning methods; then, similar experiments are conducted to show improvements made by our proposed TML-SVM over the traditional SVMs.

% \subsection{UCI Datasets}
% In this section, we evaluate the classification performance of ML--TPS on seven widely used datasets from UCI machine learning repository. The leftmost column of Table~\ref{T:tatble} summarizes the details of these datasets (i.e., the three numbers in the $[$ $]$ indicate the data size, feature dimension, and number of classes for the corresponding dataset ). All datasets have been preprocessed through normalization. 

\subsection{Comparisons with NN based ML solutions}
%The vast majority of the existing supervised metric learning solutions \cite{NCA,ITML,weinberger2009distance,LMNN2009,msNCA,wang2012parametric} were designed and can be applied to improve metric based classification methods, especially the Nearest Neighbor (NN) classifiers. Thus, 
 The first set of experiments are within the Nearest Neighbor (NN) category. We choose $k=1$ in $k$NN, and the five competing metric learning methods are: Large Margin Nearest Neighbor classification (LMNN) \cite{weinberger2009distance}, Information-Theoretic Metric Learning (ITML) \cite{ITML}, Neighborhood Components Analysis (NCA) \cite{NCA},  GB-LMNN \cite{kedem2012non} and Parametric Local Metric Learning (PLML) \cite{wang2012parametric}. %Also, we compare ML--TPS with SVM, for which the one-against-all strategy is used in multi-class classification problems. 
The hyper-parameters of NCA, ITML, LMNN and GB-LMNN are set by following \cite{NCA,ITML,weinberger2009distance,kedem2012non} respectively. PLML has a number of hyper-parameters, so we follow the suggestion of \cite{wang2012parametric}: use a 3-fold CV to select $\alpha_2$ from $\{0.01\sim1000\}$, and set the other hyper-parameters by its default. In our TML-NN model, there are two hyper-parameters: the number of anchor points $p$ and the weighting factor $\lambda$. For $p$, we empirically set it to $30\%$ of the training samples; for $\lambda$, we select it through CV from $\{5^{-5}\sim5^{25}\}$.

\begin{table}[!ht]
\small
\centering
\vspace*{-0.1in}
\caption{Mean and standard deviation of $k$NN based classification accuracy results on seven UCI datasets. Boldface denotes the highest classification accuracy for each dataset. The superscripts $^{+-=}$ in TML-NN column indicate a significant win, loss or no difference respectively based on the pairwise Student's $t$-test with the other six methods. The number in the parenthesis denotes the score of the respective method for the given dataset.}
%\vspace*{-0.1in}
\scalebox{0.79}{
\begin{tabular}{>{\columncolor{Gray}}c c c c c c c c}
\specialrule{.2em}{0em}{0em}
\rowcolor{Gray}
\multirow{1}{*}{Datasets}  & $k$NN & LMNN & ITML & NCA & PLML & GB-LMNN & TML-NN \\
%$[$\#Inst./\#Feat./\#Class$]$  
%&  &  & & & & & \\
\specialrule{.1em}{0em}{0em}
 %\multirow{-3}{*}{Iris(150/4/3)} & \multirow{-3}{*}{$95.04 \pm 3.03 $ (2.5)} & $95.71 \pm 2.94$ (3.0) & $94.77 \pm 3.00$ (2.0) & $84.38 \pm 4.91$ (0) & $95.26 \pm 3.07 $ (2.5)& $97.18 \pm 2.51^{+++++}$ \quad (5.0) \\
\multirow{1}{*}{Iris} & $95.70 \pm 2.31 $  & $95.06 \pm 2.62$  & $95.22 \pm 2.56$  & $94.68 \pm 2.35$  & $84.22 \pm 4.54 $ & $95.15 \pm 2.17$ & $\bf{96.49} \pm 2.32$\\
 $[150/4/3]$   & (3.5) & (3.0) & (3.0) & (2.5) & (0) & (3.0) & $^{++++++}$ (6.0)\\
\specialrule{.1em}{0em}{0em}
\multirow{1}{*}{Wine} & $95.21 \pm 2.04 $  & $97.25 \pm 1.80$  & $96.90 \pm 2.31$  & $96.65 \pm 2.27$  & $96.61 \pm 2.10 $ & $96.80 \pm 1.94$ & $\bf{97.28} \pm 2.07$\\
 $[178/13/3]$   & (0.0) & (4.5) & (3.5) & (2.5) & (2.5) & (3.5) & $^{+==++=}$ (4.5)\\
\specialrule{.1em}{0em}{0em}
\multirow{1}{*}{Breast} & $95.35 \pm 1.34 $  & $95.66 \pm 1.39$  & $95.76 \pm 1.30$  & $95.57 \pm 1.13$  & $\bf{96.18 \pm 0.98} $ & $96.04 \pm 1.22$ & $95.97 \pm 1.04$\\
  $[683/10/2]$  & (1.0) & (2.0) & (2.5) & (1.5) & (5.0) & (5.0) & $^{+==+==}$ (4.0)\\
\specialrule{.1em}{0em}{0em}
\multirow{1}{*}{ Diabetes} & $70.58 \pm 2.26 $  & $70.54 \pm 2.52$  & $68.81 \pm 2.65$  & $68.53 \pm 2.71$  & $69.04 \pm 2.30 $ & $70.62 \pm 2.23$ & $\bf{71.54} \pm 2.21$\\
  $[768/8/2]$  & (4.0) & (4.0) & (1.0) & (1.0) & (1.0) & (4.0) & $^{++++++}$ (6.0)\\
\specialrule{.1em}{0em}{0em}
\multirow{1}{*}{Liver} & $61.20 \pm 3.96 $  & $60.79 \pm 3.54$  & $60.07 \pm 4.92$  & $62.63 \pm 4.15$  & $64.74 \pm 3.99 $ & $64.81 \pm 3.80$ & $\bf{64.97} \pm 4.28$\\
  $[345/6/2]$  & (1.0) & (1.0) & (1.0) & (3.0) & (5.0) & (5.0) & $^{++++==}$ (5.0)\\
\specialrule{.1em}{0em}{0em}
\multirow{1}{*}{Sonar} & $84.73 \pm 3.45 $  & $84.12 \pm 4.13$  & $82.14 \pm 5.94$  & $85.46 \pm 3.51$  & $\bf{87.42} \pm 4.70 $ & $85.48 \pm 4.04$ & $85.35 \pm 3.82$\\
  $[208/60/2]$  & (3.0) & (1.5) & (0) & (3.5) & (6.0) & (3.5) & $^{=++=-=}$ (3.5)\\
\specialrule{.1em}{0em}{0em}
\multirow{1}{*}{Ionosphere} & $85.83 \pm 2.62 $  & $88.40 \pm 2.54$  & $87.45 \pm 3.07$  & $88.33 \pm 2.77$  & $\bf{91.03} \pm 2.23 $ & $89.47 \pm 2.70$ & $88.79 \pm 2.37$\\
  $[351/34/2]$  & (0) & (3.0) & (1.0) & (3.0) & (6.0) & (4.5) & $^{+=+=-=}$ (3.5)\\
\specialrule{.1em}{0em}{0em}
\specialrule{.1em}{0em}{0em}
 Total Score & $12.5$ & $19.0$ & $12.0$ & $17.0$ & $25.5$ & $28.5$ & $32.5$ \\
\specialrule{.2em}{0em}{0em}
\end{tabular}
}
%\vspace{-0.05in}
\label{T:tatble}
\end{table}
%\end{table}

To better compare the classification performance, we run the experiment 100 times with different random 3-fold splits of each dataset, two for training and one for testing.  Furthermore, we conduct a pairwise Student's $t$-test with a $p$-value 0.05 among the seven methods for each dataset. Then, a ranking schema from \cite{wang2012parametric} is used to evaluate the relative performance of these algorithms: a method A will be assigned 1 point if it has a statistically significantly better accuracy than another method B; 0.5 points if there is no significant difference, and 0 point if A performs significantly worse than B. The experiment results by averaging over the 100 runs are reported in Table~\ref{T:tatble}.

From Table~\ref{T:tatble}, we can see that TML-NN outperforms the other six methods in a statistically significant manner, with a total score of $32.5$ points. Out of the total $42$ pairwise comparisons, TML-NN has $25$ statistical wins. Furthermore, it has significantly improved the baseline method, $k$NN, on six datasets out of the total seven, and performed equally well on the seventh (``Sonar''). It is also worth pointing out that the proposed nonlinear TML-NN has $14$ wins and no loss out of the total $18$ comparisons against the linear ML solutions (LMNN, ITML, NCA); against the local nonlinear ML solutions (PLML, GB-LMNN), TML-NN has five wins and only two loss out of the total $14$ comparisons.  

%The local metric learning method PLML has the worst performance, which may result from the fact that its multi-metric smoothing strategy is not very suitable for the datasets in our experiments, which have a relatively small number of classes.   

\subsection{Improvements over SVMs}  
%In our ML--TPS model, feature spaces are deformed 
%better metrics are obtained to pull same-class elements closer and push between-class further away. Although not directly designed under SVMs' paradigm, our approach has a positive effect in enlarging the separation between two classes and pushing/pulling elements to the correct class sides. In other words, if used as a preprocessing step, our model can potentially facilitate SVMs to find the maximum-margin hyperplane and therefore leads to improved performance. % Recently, Xu, Weinberger and Chapelle \cite{SVML} pointed out that metric learning methods can be used as the preprocessing step to transform the feature space for, or combined with SVMs to improve SVMs' performance. 

To verify the effectiveness of our proposed nonlinear metric learning for SVMs, we conduct  another set of experiments on the same seven UCI datasets to compare the following four SVM models: linear SVM ($\it{l}$-SVM),  kernel SVM ($\it{r}$-SVM), our proposed TML-SVM and kernel TML-SVM. For $\it{l}$-SVM, we directly utilize the off-the-shelf LIBSVM solver \cite{chang2011libsvm}, for which the slackness coefficient $C$ are tuned through 3-fold CV from $\{2^{-5}\sim 2^{15}\}$. For $\it{r}$-SVM, we choose the Gaussian kernel and select the kernel width $\sigma$ through CV from $\{d_{min}\sim 20 d_{min}\}$, where $d_{min}$ is the mean of the distances between a input data to its nearest neighbor. TML-SVM has three hyper-parameters to be tuned: the number of anchor points $p$ and the tradeoff coefficients  $C_1$ and $C_2$. For $p$, we still empirically set it to $30\%$ of the training samples; for $C_1$ and $C_2$, we select them through CV from $\{2^{-5}\sim2^{15}\}$ and $\{
5^{-5}\sim5^{25}\}$ respectively. In kernel TML-SVM, we use the same Gaussian kernel width $\sigma$ selected in $\it{r}$-SVM for each dataset, and tune the other parameters $C_1$ and $C_2$ similarly as in TML-SVM. To deal with multi-class classification, we apply the ``one-against-one'' strategy on top of binary TML-SVM and kernel TML-SVM, the same as used in LIBSVM.   
\vspace*{-0.1in}
\begin{wraptable}{l}{0.67\textwidth}
\caption{Mean and standard deviation of SVMs based classification accuracy results on seven UCI datasets. The settings and notations of the comparison scores are identical to those in Table 1.}
\vspace*{0.05in}
%The superscripts $+-=$  indicate a significant win, loss or no difference respectively based on the pairwise Student's t-test with the other three methods. The number in the parenthesis denotes the score of the respective method for the given dataset.
%\vspace*{-0.1in}
\scalebox{0.78}{
\begin{tabular}{>{\columncolor{Gray}}c c c c c }
\specialrule{.2em}{0em}{0em}
\rowcolor{Gray}
 Datasets & \it{l}-SVM & TML-SVM  & \it{r}-SVM & kernel TML-SVM  \\
\specialrule{.1em}{0em}{0em}
\multirow{1}{*}{Iris} & $95.94 \pm 2.42$ & $96.67 \pm 2.31$  & $96.09 \pm 2.34 $ & $\bf{96.81} \pm 2.43$ \\
    & $^{-=-}$ (0.5) & $^{+==}$(2.0) & $^{==-}$(1.0) & $^{+=+}$(2.5) \\
\specialrule{.1em}{0em}{0em}
\multirow{1}{*}{Wine} & $97.20 \pm 1.86$ & $\bf{98.97} \pm 1.25$  & $98.07 \pm 1.80 $ & $98.46 \pm 1.46$ \\
    & $^{---}$ (0) & $^{+++}$(3.0) & $^{+--}$(1.0) & $^{+-+}$(2.0) \\
\specialrule{.1em}{0em}{0em}
\multirow{1}{*}{Breast} & $96.73 \pm 0.97$ & $97.15 \pm 0.88$  & $97.06 \pm 0.83 $ & $\bf{97.44} \pm 0.98$ \\
    & $^{---}$ (0) & $^{+=-}$(1.5) & $^{+=-}$(1.5) & $^{+++}$(3.0) \\
\specialrule{.1em}{0em}{0em}
\multirow{1}{*}{Diabetes} & $76.66 \pm 2.18$ & $77.24 \pm 1.92$  & $77.07 \pm 2.05 $ & $\bf{77.69} \pm 2.20$ \\
    & $^{-=-}$ (0.5) & $^{+==}$(2.0) & $^{==-}$(1.0) & $^{+=+}$(2.5) \\
\specialrule{.1em}{0em}{0em}
\multirow{1}{*}{Liver} & $69.06 \pm 3.79$ & $72.62 \pm 3.13$  & $72.35 \pm 3.76 $ & $\bf{73.40} \pm 3.58$ \\
    & $^{---}$ (0) & $^{+==}$(2.0) & $^{+=-}$(1.5) & $^{+=+}$(2.5) \\
\specialrule{.1em}{0em}{0em}
\multirow{1}{*}{Sonar} & $75.78 \pm 4.16$ & $82.16 \pm 3.79$  & $84.96 \pm 4.28 $ & $\bf{86.54} \pm 3.47$ \\
    & $^{---}$ (0) & $^{+--}$(1.0) & $^{++-}$(2.0) & $^{+++}$(3.0) \\
\specialrule{.1em}{0em}{0em}
\multirow{1}{*}{Ionosphere} & $87.75 \pm 2.42$ & $92.94 \pm 2.06$  & $94.36 \pm 1.87 $ & $\bf{95.12} \pm 1.72$ \\
    & $^{---}$ (0) & $^{+--}$(1.0) & $^{++-}$(2.0) & $^{+++}$(3.0) \\
\specialrule{.16em}{0em}{0em}
%\specialrule{.1em}{0em}{0em}
 Total Score  & $1.0$ & $12.5$ & $10.0$ & $18.5$ \\
\specialrule{.2em}{0em}{0em}
\end{tabular}
}
\label{T:tatble2}
%\end{table}
\end{wraptable} 

 We adopt the same experimental setting and statistical ranking scheme as in the NN based classification, and report the results averaged from 100 runs in Table~\ref{T:tatble2}. It is evident that combining our proposed nonlinear metric learning has significantly improved the performance of both $\it{l}$-SVM and $\it{r}$-SVM. To be specific, TML-SVM outperforms $\it{l}$-SVM on all seven datasets; kernel TML-SVM also fares better than $\it{r}$-SVM on all seven datasets. %does better than $\it{r}$-SVM on four datasets (``Iris'', ``Breast'', ``Diabetes'', and ``Ionosphere''), and comparably well on the other three. 
Furthermore, it is worth pointing out that TML-SVM has significantly improved $\it{l}$-SVM's classification rates, performing better than or comparable to $\it{r}$-SVM on five datasets (``Iris'', ``Wine'', ``Breast'', ``Diabetes'', and ``Liver'').

\section{Conclusion}

In this paper, we present two nonlinear metric learning solutions, for $k$NN and SVMs respectively, based on geometric transformations. The novelty of our approaches lies in the fact that it generalizes the linear or piecewise linear transformations in traditional metric learning solutions to a globally smooth nonlinear deformation in the input space. The geometric model used in this paper is thin-plate splines, and it can be extended to other radial distance functions. To explore other types of geometric models from the perspective of conditionally positive definite kernels is the direction of our future efforts. We are also interested in investigating a more efficient numerical optimization scheme (or the analytic form) for the proposed TPS based methods.

\small{
\bibliographystyle{unsrt}
\bibliography{nips2015}
}
\end{document}